%% file: main.tex
\documentclass[letterpaper, 10 pt, conference]{ieeeconf}

\IEEEoverridecommandlockouts
\usepackage{amsmath, amssymb}

\usepackage{graphics} 
\usepackage{epsfig} 
\usepackage{times} 
\usepackage{amsmath} 
\usepackage{amssymb}  
\usepackage[table]{xcolor}
\usepackage[ruled,linesnumbered, noend]{algorithm2e}
\SetKwRepeat{Do}{do}{while}
\usepackage{adjustbox}
\usepackage{wrapfig}

\usepackage{soul}
\usepackage{booktabs}

\usepackage{cite}
\usepackage{comment}
\usepackage{multirow}
\usepackage{graphicx}
\usepackage{wrapfig}
\usepackage{balance}
\usepackage{subfigure}
\usepackage{capt-of}

\newcommand{\ve}[1]{\mathbf{#1}} 
\newcommand{\hve}[1]{\hat{\mathbf{#1}}} 
\newcommand{\dve}[1]{\dot{\mathbf{#1}}} 

\usepackage{color, colortbl}
\definecolor{Gray}{gray}{0.9}

\usepackage{pifont}
\title{Vibration-based Full State In-Hand Manipulation of Thin Objects}

\author{Oron Binyamin, Guy Shapira, Noam Nahum and Avishai Sintov
\thanks{O. Binyamin, G. Shapira, N. Nahum and  A. Sintov are with the School of Mechanical Engineering, Tel-Aviv University, Israel. Corresponding Author: sintov1@tauex.tau.ac.il.}
}

\begin{document}


\maketitle

\begin{abstract}
Robotic hands offer advanced manipulation capabilities, while their complexity and cost often limit their real-world applications. In contrast, simple parallel grippers, though affordable, are restricted to basic tasks like pick-and-place. Recently, a vibration-based mechanism was proposed to augment parallel grippers and enable in-hand manipulation capabilities for thin objects. By utilizing the stick-slip phenomenon, a simple controller was able to drive a grasped object to a desired position. However, due to the underactuated nature of the mechanism, direct control of the object's orientation was not possible. In this letter, we address the challenge of manipulating the entire state of the object. Hence, we present the excitation of a cyclic phenomenon where the object's center-of-mass rotates in a constant radius about the grasping point. With this cyclic motion, we propose an algorithm for manipulating the object to desired states. In addition to a full analytical analysis of the cyclic phenomenon, we propose the use of duty cycle modulation in operating the vibration actuator to provide more accurate manipulation. Finite element analysis, experiments and task demonstrations validate the proposed algorithm. 
\end{abstract}


\section{Introduction}
\label{sec:introduction}
\input{introduction}

\section{System \& Method}
\label{sec:System}
\input{system}

\section{Experiments}
\label{sec:experiments}
\input{experiments}

\section{Conclusions}
\input{conclusions}

\bibliographystyle{IEEEtran}
\bibliography{ref}

\end{document}

%% file: introduction.tex
In-hand manipulation typically requires either a dexterous robotic hand with multiple degrees of freedom (DOF) \cite{Weinberg2024} or complex manipulation strategies involving dynamic interactions with the environment \cite{SintovSwingUp2016, Cruciani2018} or environmental engagement \cite{Dafle2014,Dafle2020}. However, these approaches often demand sophisticated sensing and control systems \cite{Hertkorn2013}, leading to increased complexity and cost. Such factors can hinder the practicality of these methods in applications like assembly lines or medical procedures.
\begin{figure}[h]
    \centering
    \includegraphics[width=\linewidth]{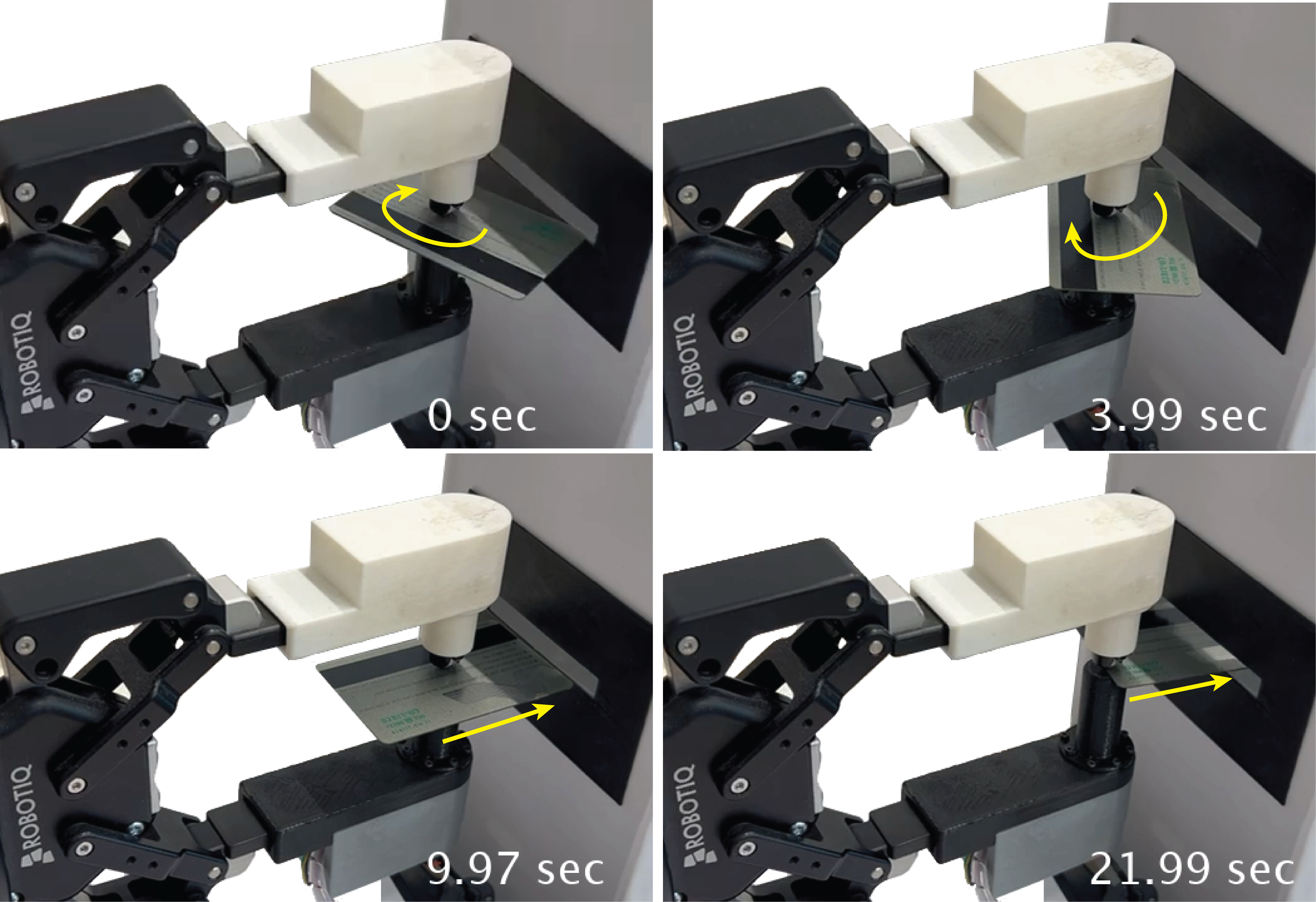} 
    \caption{A parallel gripper equipped with the Vibration Finger Manipulator (VFM) demonstrates the manipulation of a credit card into an ATM-like slot. Using only vibration, the card is first aligned with the slot through rotational motion, followed by linear motion into the slot.}
    \label{fig:front}
\end{figure}

Parallel jaw grippers are a class of low DOF end-effectors that are unable to perform intrinsic in-hand manipulations. They are inherently limited to a single DOF, restricting their ability to manipulate objects within their grasp. Nevertheless, they are renowned for their simplicity, durability and affordability. These are widely adopted in industrial applications. Their versatility allows for precise grasping of various objects, making them indispensable in material handling tasks \cite{Guo2017}. The prevalent manipulation strategy for these grippers involves a pick-and-place approach, where objects are placed on some surface and re-picked with a different grasp configuration \cite{Zeng2018}. However, this method can be time-consuming and requires ample workspace, limiting its applicability in certain scenarios. Several mechanisms have been proposed to augment the capabilities of parallel jaw grippers such as an active conveyor surface \cite{Ma2016} and pneumatic braking mechanism \cite{Taylor2020}. However, these additions often introduce complexity and may be limited to specific manipulation tasks \cite{Terasaki1998,Zuo2021,Chapman2021}.

Vibration has been employed as a mechanism for object manipulation since the pioneering work of Chladni with horizontally vibrating plates \cite{chladni1787} and the establishment of vibrating conveyor belts in industrial part management \cite{Winkler1978}. These have contributed to the existing technology of vibrating systems often based on acoustic \cite{Zhou2016,Latifi2017} or mechanical excitation \cite{Bohringer1995, Reznik1998, Breguet1998, Du1999, Mayyas2020,Kopitca2021}. A fundamental aspect of vibration-based manipulation is the \textit{Stick-Slip} effect \cite{Gao1994}. This phenomenon involves alternating between static friction (no relative motion) and kinetic friction (sliding motion) at the contact interface between two surfaces. More focused toward robotic applications, the Stick-Slip effect was harnessed to design simple, low-cost micro-robots. A notable example is the use of two collinear vibration motors, as proposed in \cite{Vartholomeos2006}, to achieve effective locomotion. This approach inspired the development of the Kilobot platform, a widely used tool in swarm robotics research \cite{Rubenstein2014,Rubenstein2014b}.

\begin{figure*}
    \centering
    \includegraphics[width=\linewidth]{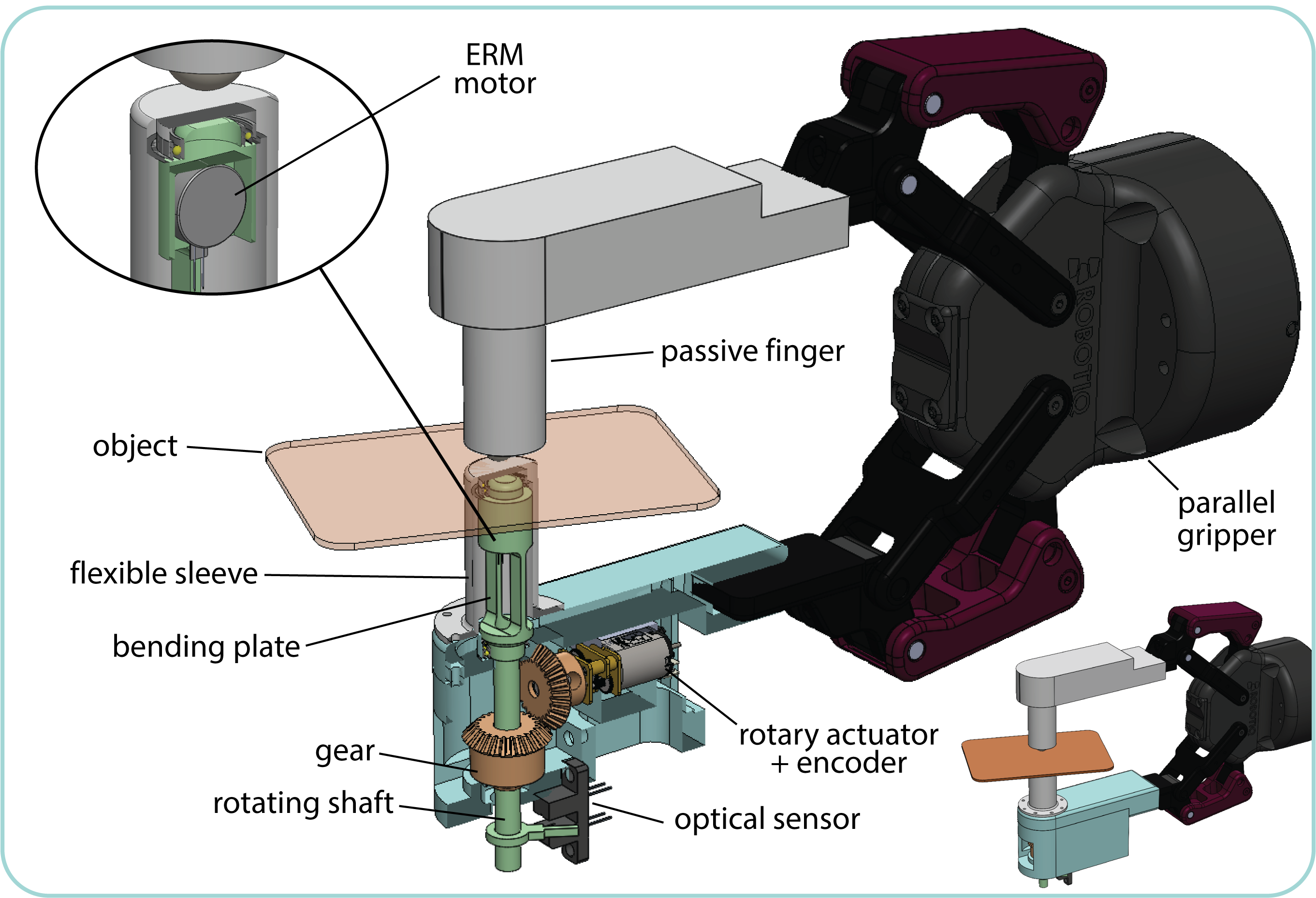} 
    \caption{Design of the Vibratory Finger Manipulator (VFM) and its mount on a parallel jaw gripper.}
    \label{fig:CAD}
\end{figure*}

Vibration has been primarily used in robotic hands to control slippage, without additional manipulation capabilities \cite{Rong2014, Long2017, Gao2019, Suzuki2021}. Recently, vibration was introduced to in-hand manipulation by parallel jaw grippers \cite{Nahum2022}. A novel mechanism termed the \textit{Vibratory Finger Manipulator} (VFM) was proposed where an off-the-shelve vibration motor within one jaw manipulates a grasped object. A simple rotary actuator allows for precise control of the vibration direction, enabling precise position manipulation of the grasped object. While the proposed controller demonstrated partial stability \cite{Vorotnikov1995} with accurate position control, it was unable to control the object's orientation. Furthermore, the effect of vibration frequency on the control's performance was not considered. 

In this letter, we address the manipulation challenge of a thin object over its full state using a VFM-based parallel gripper. The VFM-object system is underactuated and does not enable direct control over the entire state and, particularly, the orientation. Hence, we decouple the position and orientation components of the manipulation task and propose a manipulation algorithm. Through rigorous analysis of the VFM-object system, we devise an approach to exert cyclic motion of the object to reach the desired orientation. Then, linear motion through the object's center-of-mass will enable rotation-free motion to the desired position. In addition, we explore the use of duty cycle modulation to operate the vibration motor over the simple application of a constant voltage. We show that the duty cycle better maintains the orientation of the object, yielding increased accuracy. The proposed mechanism and vibration-based control have potential applications in fields such as medical procedures, requiring precise manipulation of delicate instruments like surgical knives. Additionally, it could be utilized for tasks like textile and tissue handling, plastic card (e.g., security and ATM cards) insertion as demonstrated in Figure \ref{fig:front} or manipulating keys.

%% file: system.tex
\begin{figure*}[h]
    \centering
    \begin{tabular}{cc}
        \includegraphics[height=5cm]{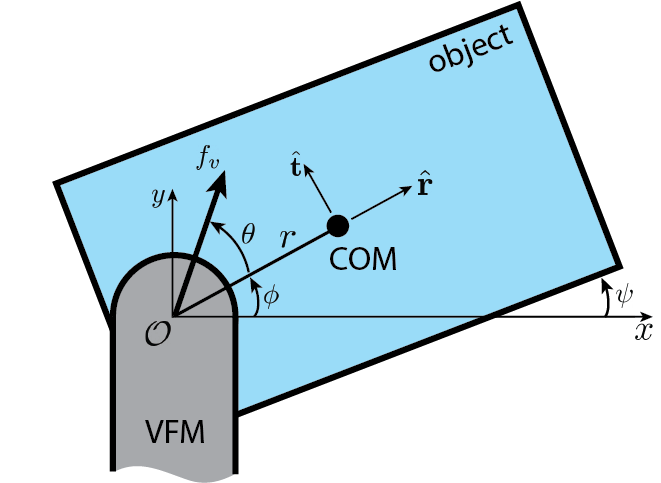} & \includegraphics[height=5cm]{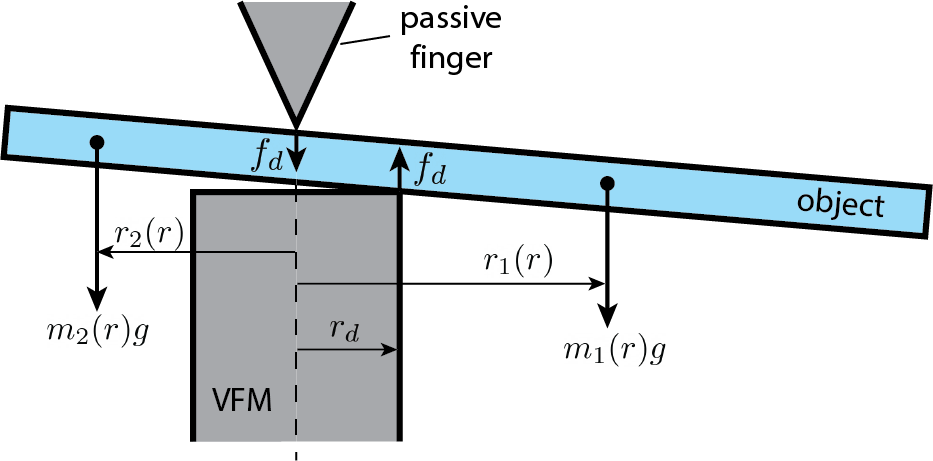} \\
        (a) & (b)
    \end{tabular}
    \caption{Illustration of (a) a bottom-view where vibration force $f_v$ is exerted onto the object by the VFM and (b) a side view of the object's tilt due to gravity.}
    \label{fig:finger_object}
\end{figure*}

\subsection{Design}
\label{sec:design}

The Vibration Force Module (VFM) was proposed by Nahum and Sintov \cite{Nahum2022} for in-hand object manipulation of thin objects within a parallel gripper. We briefly present its design mechanism seen in Figure \ref{fig:CAD}. The VFM comprises an Eccentric Rotating Mass (ERM) motor and a rotary actuator. The ERM motor, housed within a 3D-printed shaft, generates vibrations perpendicular to the shaft axis. The shaft, made with Polylactic-Acid (PLA) filament, is supported by three bearings to ensure concentric rotation with minimal friction. The shaft also includes a thin bending plate that amplifies the vibrations by having its normal perpendicular to the ERM axis and, thus, minimizing energy loss. The entire assembly is encased in a flexible Thermoplastic polyurethane (TPU) sleeve to enhance vibration transmission and protect internal components.

The rotary actuator, equipped with an encoder, controls the shaft's orientation, enabling directional vibration control. An optical sensor enables the calibration of the encoder and rotating shaft. In addition, a passive finger equipped with a roller ball bearing opposes the vibrating finger, creating an initial gripping force. Both fingers exert an initial normal force $f_b$ on the object to establish a grasp. This mechanism induces the \textit{Stick-Slip} effect (detailed below), facilitating object manipulation without external moving parts. 


\subsection{Model}

\subsubsection{Object State Definition}

Let the system's coordinate frame $\mathcal{O}$ be at the center of the VFM. 
We define the position of the Center-Of-Mass (COM) of an object with mass $M$ at time $t$ in polar coordinates $\ve{x}(t)=(r(t),\phi(t))$ as illustrated in Figure \ref{fig:finger_object}a. Hence, its linear velocity and acceleration are given by
\begin{align}
    \ve{v}(t) &= \dot{r}\hve{r}+r\dot{\phi}\hve{t} \\
    \ve{a}(t) &= (\ddot{r}-r\dot{\phi}^2)\hve{r}+(2\dot{r}\dot{\phi}+r\ddot{\phi})\hve{t} 
\end{align}
where $\hve{r}=(\cos\phi, \sin\phi)^T$ and $\hve{t}=(-\sin\phi, \cos\phi)^T$ are the radial and tangential unit vectors, respectively. The orientation of the object is denoted by $\psi$. Hence, the state of the object is defined by vector $\ve{s}=(r,\phi,\psi)^T$.


\subsubsection{Vibration Movement Forces}

The small eccentric mass $m$ on a link of length $l$ within the ERM rotates in a constant frequency $\omega$. For our analysis and for brevity, we assume that the gripper's grasping axis is vertical to the ground having the object nearly horizontal. Nevertheless, 
prior work has shown the ability of the VFM to manipulate an object while inclined \cite{Nahum2022}. In this configuration within the VFM, the tangential and normal forces exerted on the object due to the vibration are 
\begin{equation}
    f_v(t) = m l \omega^2\cos(\omega t)
    \label{eq:f_v}
\end{equation}
and 
\begin{equation}
    f_n(t) = m l \omega^2\sin(\omega t),
    \label{eq:f_n}
\end{equation}
respectively. The direction of force $f_v(t)$ is defined by steering angle $\theta(t)$ relative to unit vector $\hve{r}$. Angle $\theta(t)$ is controlled by the rotary actuator and encoder.

The net normal force exerted on the object is 
\begin{equation}
    f_N(t) = f_b + f_n(t) + Mg + f_d(r)
    \label{eq:f_N}
\end{equation}
where $g$ is the gravitational acceleration. Force $f_d$ is part of a force couple required to balance the tilt of the object due to gravity as seen in Figure \ref{fig:finger_object}b. This would occur when the gripping point is not at the object's COM. With the assumption of a small tilt angle, the equilibrium net torque about the longitude axis of the VFM yields
\begin{equation}
    f_d(r) = \frac{g}{r_d}\left( m_1(r)r_1(r)-m_2(r)r_2(r) \right)
    \label{eq:f_d}
\end{equation}
where $r_d$ is the radius of the vibrating finger and $m_i(r)$ is the mass of side $i\in\{1,2\}$ of the object with COM at $r_i(r)$ such that $m_1(r)>m_2(r)$. Both quantities are a function of $r$ and depend on the specific geometry of the object. However, when $m_1(r)$ and $r_1(r)$ increase, then $m_2(r)$ and $r_2(r)$ decrease, and vice versa.

Forces $f_v(t)$ and $f_N(t)$ are sinusoidal and their relative magnitudes over time $t$ will synchronize the slip-stick transition. Let $\mu_s$ and $\mu_k$ be the static and kinetic coefficients of friction, respectively, between the finger and object. If $|f_v|\leq\mu_s|f_N|$, then the system is in the stick mode with static friction. On the other hand, the slip mode occurs when 
\begin{equation}
    |f_v|>\mu_s|f_N|
    \label{eq:fv>fk}
\end{equation}
leading to an opposing kinetic friction force 
\begin{equation}
    f_k(t) = \mu_k f_N(t).
    \label{eq:f_k}
\end{equation}
From \eqref{eq:fv>fk}, we derive the following condition for slip mode to occur:
\begin{equation}
    \label{eq:motion_constraint}
    ml\omega^2\left|\cos(\omega t)-\mu_s\sin(\omega t)\right| - \mu_s(Mg + f_d + f_b) > 0. 
\end{equation}
To satisfy \eqref{eq:motion_constraint}, one can either sufficiently increase vibration frequency $\omega$, reduce clamping force $f_b$ or reduce the tilt force $f_d$. Reducing $f_b$ often requires a highly sensitive force controlled gripper which is not always available and usually can be only coarsely modified. However, reducing $f_d$ is also possible by decreasing $r$, i.e., moving toward the object's COM, but may contradict the desired motion.

\subsubsection{Dynamic Model}

With the above forces, the following equations governs the motion of the object:
\begin{align}
    M(\ddot{r}-r\dot{\phi}^2) &= (f_v-f_k)\cos\theta \label{eq:eq_r}\\
    M(2\dot{r}\dot{\phi}+r\ddot{\phi}) &= (f_v-f_k)\sin\theta \label{eq:eq_t}\\
    I\ddot{\psi} &= r(f_v-f_k)\sin\theta \label{eq:eq_psi} 
\end{align}
We assume that the torsional friction \cite{Sintov2016tro} at the contact point is negligible and later validate this in the experiments. Eqs. \eqref{eq:eq_r} and \eqref{eq:eq_t} are acquired by summing the forces in the $\hve{r}$ and $\hve{t}$ directions, respectively. Eq. \eqref{eq:eq_psi} is the net torque about the axis perpendicular to the surface of the object. Although $f_v$ and $f_k$ are sinusoidal, we assume that the net force $f_v-f_k$ exerted on the object is constant and defined by the frequency $\omega$. Hence, force $f_v$ pushes the object in direction defined by steering angle $\theta$ with the resistance of friction force $f_k$.

By enforcing $\theta=0$ or $\theta=\pm\pi$ starting from time $t_0$, Eq. \eqref{eq:eq_psi} will yield 
\begin{equation}
    \label{eq:ddpsi}
    I\ddot{\psi}(t)=0
\end{equation}
for $t>t_0$. If the initial angular velocity is zero, i.e., $\dot{\psi}(t_0)=0$, then the motion will be solely along the $\hve{r}$ axis with no rotational change of the object. This observation was used in prior work to move the position of the COM while maintaining a constant rotation angle $\psi$ \cite{Nahum2022}. By choosing either $\theta=\pm\pi$ or $\theta=0$, it is possible to move the COM toward or away from $\mathcal{O}$, respectively, without varying $\psi$.

\subsubsection{Object Rotation}
\label{sec:obj_rotate}

The VFM mechanism is underactuated and cannot directly control both position and orientation of the object. Hence, to modify the orientation of the object, we propose to exert cyclic motion on the object to generate rotation. That is, we enforce motion of the object's COM in a constant radius $r(t)=r_c$ such that $\dot{r}(t)=\ddot{r}(t)=0$. In such a case, Equations \eqref{eq:eq_r}-\eqref{eq:eq_psi} will be updated to
\begin{align}
    -Mr_c\dot{\phi}^2 &= (f_v-f_k)\cos\theta \label{eq:eq_r_rd}\\
    Mr_c\ddot{\phi} &= (f_v-f_k)\sin\theta \label{eq:eq_t_rd}\\
    I\ddot{\psi} &= r_c(f_v-f_k)\sin\theta. \label{eq:eq_psi_rd} 
\end{align}
By dividing \eqref{eq:eq_t_rd} by \eqref{eq:eq_r_rd}, we acquire the rule
\begin{equation}
    \label{eq:tantheta}
    \tan\theta=-\frac{\ddot{\phi}}{\dot{\phi}^2}.
\end{equation}
for the required steering angle $\theta$ with respect to $\dot{\phi}$ and $\ddot{\phi}$. As expected, Eq. \eqref{eq:tantheta} constrains a non-zero angular velocity $\dot{\phi}(t)\neq0$. However, measuring $\dot{\phi}$ and $\ddot{\phi}$ in real-time is usually infeasible. Hence, we consider a specific case where a steering angle $\theta=\pm\pi$ is set at time $t_\pi$. Consequently, we acquire from \eqref{eq:eq_t_rd} that the angular acceleration of $\phi$ is $\ddot{\phi}=0$. From \eqref{eq:eq_r_rd}, we get
\begin{equation}
    \label{eq:phi^2}
    \dot{\phi}^2=\frac{(f_v-f_k)}{Mr_c}.
\end{equation}
Eq. \eqref{eq:phi^2} implies that $r_c$ must be non-zero to exert circular motion. That is, circular motion about the COM with $r(t)=0$ is impossible.

When choosing $\theta=\pm\pi$ on \eqref{eq:eq_psi_rd} we get $\ddot{\psi}(t)=0$ as in \eqref{eq:ddpsi}. As stated previously, having no initial velocity (i.e., $\dot{\psi}(t_\pi)=0$) will yield motion of the COM along the $\hve{r}$-axis with no object orientation change. However, if the initial angular velocity is non-zero $|\dot{\psi}(t_\pi)|>0$ in the integration of \eqref{eq:ddpsi}, it must be that $\dot{\psi}(t)$ is constant for $t>t_\pi$. Hence, by applying the vibration force $f_v$ with angle $\theta=\pm\pi$ at time $t_\pi$ while $|\dot{\psi}(t_\pi)|>0$ and $r(t_\pi)=r_c>0$, the COM will move on a circular path centered at $\mathcal{O}$ with radius $r_c$ while $|\dot{\psi}(t)|>0$.

\subsection{Motion Algorithm and Control}

\subsubsection{Problem}

An object is held by the VFM and passive finger at some state $\ve{s}(0)$ with velocity $\dve{s}(0)=0$. We consider the problem of manipulating the object to a desired goal state $\ve{s}(t_g)=\ve{s}_g$ at some time $t_g$ such that $\dve{s}(t_g)=0$.

\subsubsection{Position Control}

Previous work \cite{Nahum2022} have shown that the controller
\begin{equation}
    \label{eq:controller}
    \theta(t) = \text{atan2}\left( k_2(t), k_1(t) \right) - \phi(t),
\end{equation}
with
\begin{align}
    k_1(t) &= r_g\cos\phi_g-r(t)\cos\phi(t) \\
    k_2(t) &= r_g\sin\phi_g-r(t)\sin\phi(t),
\end{align}
is partially stable \cite{Vorotnikov1995}. That is, controller \eqref{eq:controller} will drive the system to position $(r_g,\phi_g)$ but cannot control the rotation angle $\psi(t)$ to reach $\psi_g$. However, from \eqref{eq:ddpsi} with $\dot{\psi}(0)=0$, moving through the COM maintains a constant angle $\psi(t)=0$. Therefore, rotation will not occur if the object moves on a line connecting the COM and the goal position using \eqref{eq:controller}. In such case, motion is exerted toward or away from the COM, no torque is applied on the object and, therefore, the object will not rotate. To also ensure accurate control of the orientation angle $\psi$, we propose a novel algorithm detailed in the subsequent section.



\subsubsection{Frequency Management}

The normal operation of a vibration motor is to supply it with some constant voltage, leading to some frequency $\omega_o$. We propose controlling the vibration motor using duty cycle modulation. When a vibration mechanism operates with some duty cycle percentage, it introduces periodic pauses between pulses, allowing the object to settle and re-stabilize before the next pulse. This approach enhances control, particularly in stick-slip mechanisms, which rely on static friction. 
Using pulsed vibration in duty cycle control, we can effectively modulate the sticking and sliding phases of the object's motion. In continuous vibration, on the other hand, static friction is continuously broken, which can lead to sustained movements that may displace the object from its intended position, creating less controlled motion. Therefore, we argue that duty cycle control can improve the orientation maintenance of the object during translation toward the target position. We denote the non-continuous duty cycle frequency function as $\Omega(t)$. 


\subsubsection{Algorithm}

In the proposed algorithm, the object is moved to the origin $\mathcal{O}$, rotated with some small radius $r_c$ to $\psi_g$ and then moved along $\hve{r}$ to $r_g$. That is, rotation and position are dealt sequentially. Algorithm \ref{alg:main} describes the sequence of operations to reach the goal $\ve{s}_g$. First, in Lines \ref{ln:start}-\ref{ln:finish2COM}, the object is moved along $\hve{r}$ to the origin with steering angle $\theta=\pm\pi$ based on \eqref{eq:controller}. Generally, when moving linearly along $\pm\hve{r}$, the motion is first halted by setting the vibration frequency $\omega=0$. Hence, the orientation of the object will remain constant. 

Once at the COM, rotation of the object will be initiated as described in Section \ref{sec:obj_rotate}. First, the radius $r(t)$ is increased slightly to some value $r_c$ (Line \ref{ln:rc}). Then, by applying $\theta(t)=\theta_{\Delta t}$ for a short period of time $\Delta t$, we initiate angular velocity of the object $|\dot{\psi}(t)|>0$ (Line \ref{ln:dt}). Angle $\theta_{\Delta t}\notin\{0,\pi\}$ is a pre-defined arbitrary value. After $\Delta t$ time, $\theta$ is instantly rotated to $\pi$ such that the object rotates with frequency control until reaching $\psi_g$ (Lines \ref{ln:start_rotate}-\ref{ln:end_rotate}). Motion is stopped once reached desired angle with error smaller than a pre-defined bound $\epsilon_\psi$. 

To maintain the angle, the object is returned to the COM in Lines \ref{ln:return_COM_start}-\ref{ln:return_COM_end}. This is an important step since going directly to the goal position will not be along the $\hve{r}$-axis and rotation will occur. After reaching the COM with error smaller than a pre-defined bound $\epsilon_r$, motion along axis $\hve{r}$ is initiated until reaching $(r_g,\phi_g)$ as stated in Lines \ref{ln:reach_goal_start}-\ref{ln:reach_goal_end}. However, $\theta(t)$ is defined with respect to $\hve{r}$ while $r(t)=0$. Thus, motion is initiated toward the goal with angle $\theta(t)=\phi_g$ followed by control signal $\theta(t)=0$ from \eqref{eq:controller}. Although $\theta(t)=0$ seems like a constant control signal, it is in fact a feedback control since $\theta(t)$ is a relative angle to $\hve{r}$. Any undesired change in $r(t)$ or $\phi(t)$ will enforce the change in the absolute angle of the vibration actuator. Here also, motion will finalize if the position error is smaller that $\epsilon_r$. Since the duty cycle control is better in maintaining orientation, it is used only in the translational steps (Lines \ref{ln:finish2COM}, \ref{ln:Omega1} and \ref{ln:reach_goal_end}). In the rotational motions, on the other hand, we require orientation change and use the constant frequency $\omega_o$ (Line \ref{ln:rc}). 

\begin{algorithm}
    \caption{}
    \label{alg:main}
    \SetAlgoLined
    \SetKwInOut{Input}{Input}
    \Input{Goal state $\ve{s}_g=(r_g,\phi_g,\psi_g)^T$}
    $\omega\leftarrow0$\; \label{ln:start}
    \While{$r(t) > \epsilon_r$}{ 
    $\theta(t)\leftarrow\pm\pi$ \tcc*{move to COM}
    $\omega\leftarrow \Omega(t)$\; \label{ln:finish2COM}
    }
    $\omega\leftarrow0$,~$\theta(t)\leftarrow0$\; 
    $\omega\leftarrow\omega_o$ \tcc*{initiate $r(t)=r_c$} \label{ln:rc}
    Apply $\theta(t)\leftarrow\theta_{\Delta t}$ for $\Delta t$ time\; \label{ln:dt}
    \Do{$|\psi(t)-\psi_g| > \epsilon_\psi$ \label{ln:end_rotate}}{
        $\theta(t)\leftarrow\pi$ \tcc*{rotate to $\psi_g$} \label{ln:start_rotate}
        $\omega\leftarrow\omega_o$\; 
        }
    $\omega\leftarrow0$\; 
    \While{$r(t) > \epsilon_r$ \label{ln:return_COM_start}}{
    $\theta(t)\leftarrow\pm\pi$ \tcc*{return to COM}
    $\omega\leftarrow \Omega(t)$\; \label{ln:Omega1}
    }
    $\omega\leftarrow0$\; \label{ln:return_COM_end}
    
    \While{$|r(t) - r_g|>\epsilon_r$ \label{ln:reach_goal_start}}{
        \If{$r(t)=0$}{
            $\theta(t)\leftarrow \phi_g$\;
        }
        \Else{
            $\theta(t)\leftarrow0$\;
        }
        $\omega\leftarrow \Omega(t)$ \tcc*{move to $(r_g,\phi_g)$} \label{ln:reach_goal_end}
    }
    $\omega\leftarrow0$ \tcc*{stop at goal}
\end{algorithm}


%% file: experiments.tex
\begin{figure}
    \centering
    \includegraphics[width=\linewidth]{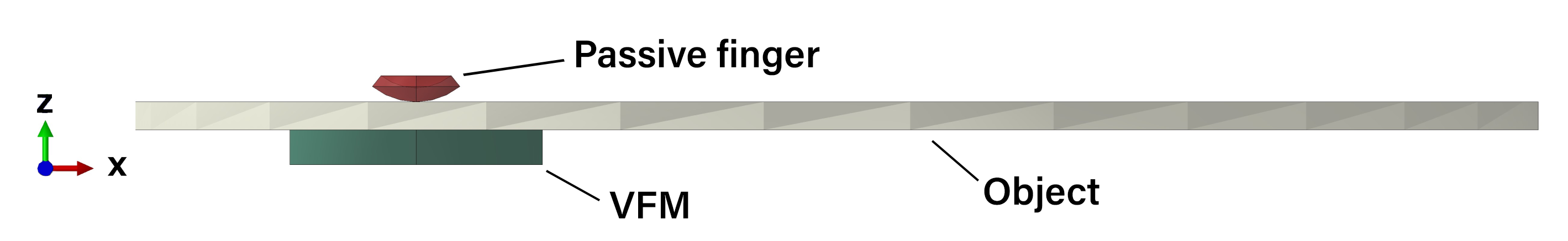} 
    \caption{The simulated FEM in Abaqus CAE-Explicit.}
    \label{fig:base_model}
\end{figure}
\begin{figure}
    \centering
    \includegraphics[width=0.9\linewidth]{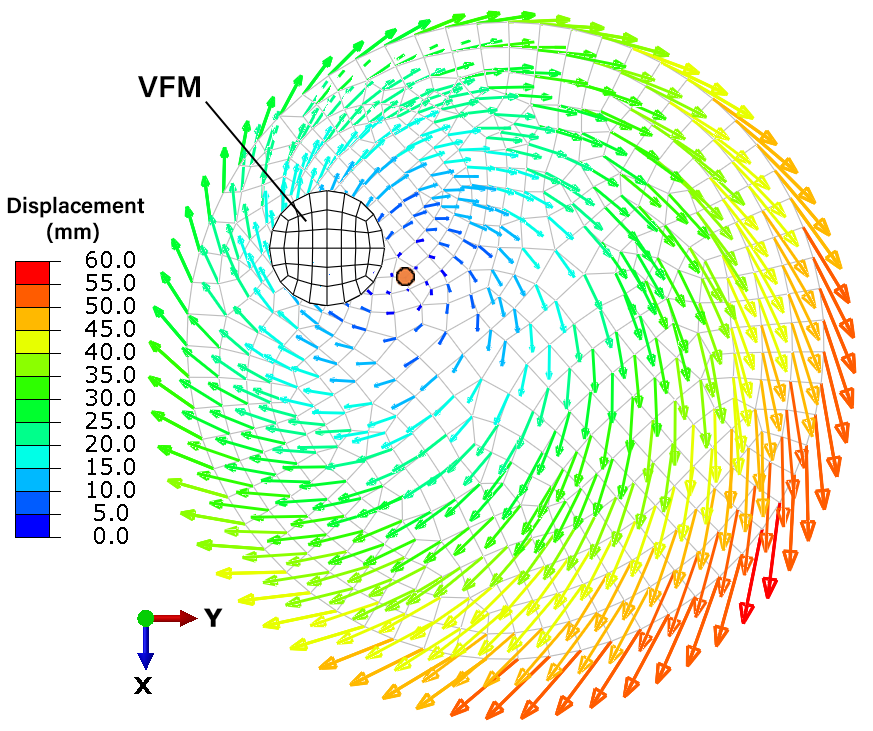} 
    \caption{FEM simulation with object displacement over 1.5 seconds of vibration excitation using the VFM. The orange circle marks the center of object rotation.}
    \label{fig:FE_Model}
\end{figure}

\begin{figure}
    \centering
    \includegraphics[width=\linewidth]{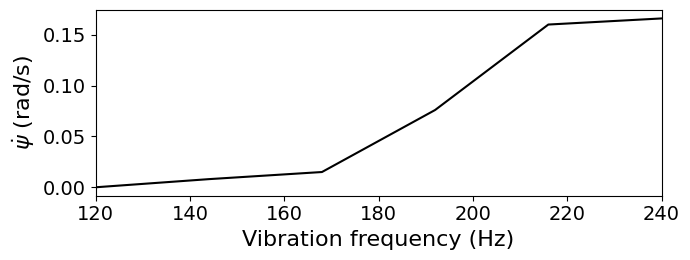} 
    \caption{Object rotation velocity $\dot{\psi}$ with respect to vibration frequency, acquired with an FEM model.}
    \label{fig:Rot_vs_freq}
\end{figure}

A prototype VFM was built using a $10\times3.4$mm Pololu ERM vibration motor, as described in Section \ref{sec:design}. 
While the VFM was fixed onto the fingers of a Robotiq 2F-85 parallel gripper, the VFM's design is compatible with various parallel gripper mechanisms. The Robotiq 2F-85 gripper was mounted on a stationary frame. To track an object's state $\ve{s}$ in real-time, a camera was positioned above the experimental setup. ArUco markers, attached to the object, were tracked at a frequency of 60 Hz, providing position and orientation data with an estimated error of approximately 1.5 mm and 1 degree, respectively. We next present Finite Element Model (FEM) analysis of the proposed algorithm and experiments with different objects. Videos of the experiments can be seen in the supplementary material.

\begin{figure*}
    \includegraphics[width=\linewidth]{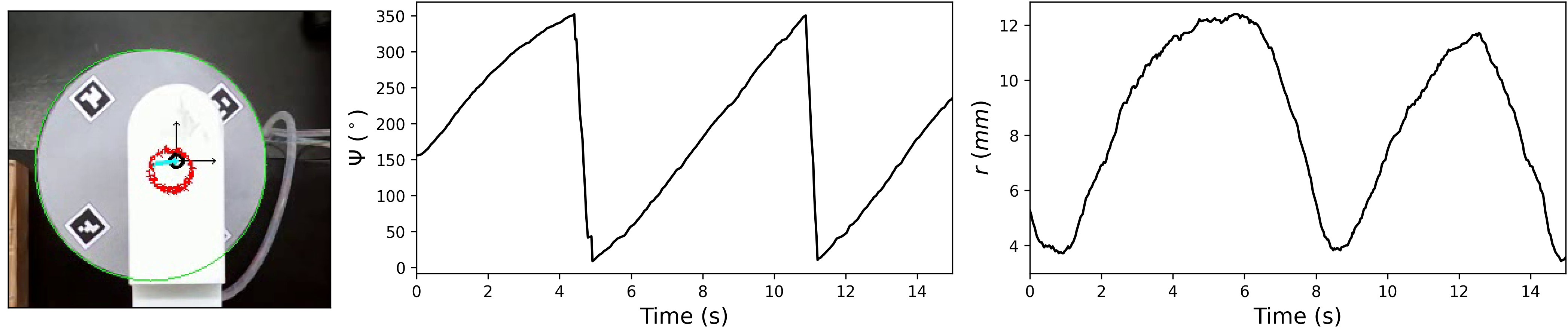} 
    \vspace{-0.8cm}
    \caption{Demonstration of an initiated cyclic motion (left) with a disc, where the curve is the path of its COM. Measured (middle) orientation angle and (right) cyclic radius are shown with respect to time.}
    \label{fig:circle}
\end{figure*}
\begin{figure*}[h]
    \includegraphics[width=\linewidth]{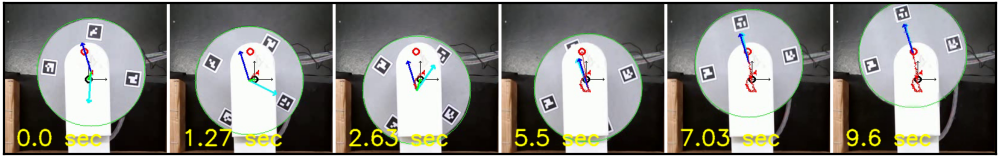} 
    \vspace{-0.8cm}
    \caption{Manipulation example for moving the object to a desired state. First, the object is rotated from its current orientation angle (cyan arrow) to a desired orientation (blue arrow). Then, it is moved through the origin to the desired position (red circle) while maintaining a constant orientation angle.}
    \label{fig:traj_snapshots}
\end{figure*}

\subsection{Finite Element Analysis}

A FEM was developed using Abaqus CAE-Explicit to validate the above mathematical model for the vibration force exerted on the object. The model facilitates the qualitative demonstration of stick-slip motion and provides insights into the influence of various parameters on the motion of the manipulated object. The model replicates physical experiments conducted using a full 3D representation of the system. The modeled system consists of three bodies illustrated in Figure \ref{fig:base_model}: a manipulated object, a VFM and a top passive finger. The manipulated object is considered a rigid body due to negligible elastic deformation, while the vibrating finger and top passive finger were modeled as elastic bodies with material properties of the TPU and steel, respectively. This approach allowed for the simulation of mechanical stiffness and the accurate calculation of contact pressure between the vibrating finger and the manipulated object. Gravitational forces were applied to the COM of the object, and an initial gripping force of 3 N was imposed between the gripping fingers. Sinusoidal vibration forces were applied according to \eqref{eq:f_v}-\eqref{eq:f_n} at the contact surface of the VFM. 

The object was exerted with the VFM to initiate object rotation according to Algorithm \ref{alg:main} without linear motion. Figure \ref{fig:FE_Model} presents the simulated displacement of the object. The displacement demonstrates cyclic motion and rotation of the object. Figure \ref{fig:Rot_vs_freq} shows the object's rotational velocity with respect to the vibration frequency, demonstrating the ability to regulate the velocity with frequency control. These results provide qualitative insights into the dynamic behavior of the system, rather than serving as a quantitative tool for precise motion prediction. Nonetheless, these simulations lay the groundwork for the following experiments on a real system.

\begin{table}[]
\centering
\caption{Motion accuracy for Algorithm \ref{alg:main}}
\label{tb:algorithm_accuracy}
\begin{tabular}{llcc}\toprule
                       &          & Position    & Orientation       \\
                       &          & error (mm)  & error ($^\circ$)  \\\midrule
& w/ frequency $\omega_o$     & \cellcolor[HTML]{C0C0C0} & 3.8 $\pm$ 1.2  \\
\multirow{-2}{*}{Disk} & w/ duty cycle $\Omega(t)$   & \multirow{-2}{*}{\cellcolor[HTML]{C0C0C0}1.88 $\pm$ 0.87} & \cellcolor[HTML]{C0C0C0}1.5 $\pm$ 1.02   \\ 
\hline
\multirow{2}{*}{Rec.}  &  w/ frequency $\omega_o$ & \cellcolor[HTML]{C0C0C0} & 4.5 $\pm$ 1.5 \\
  &  w/ duty cycle $\Omega(t)$ & \multirow{-2}{*}{\cellcolor[HTML]{C0C0C0}1.92 $\pm$ 0.90}                  & \cellcolor[HTML]{C0C0C0}2.7 +- 1.3 \\
\bottomrule
\end{tabular}
\end{table}

\begin{figure}[h]
    \includegraphics[width=\linewidth]{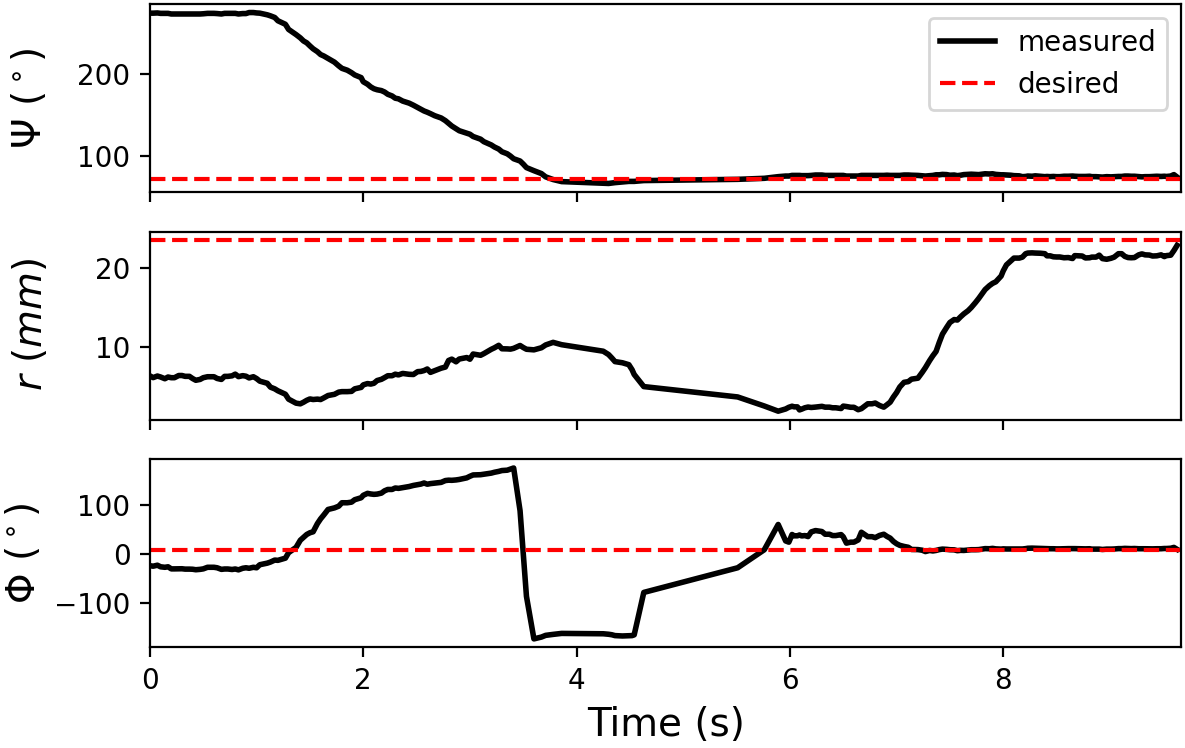} 
    \caption{State trajectory of the manipulation example seen in Figure \ref{fig:traj_snapshots}, with respect to time. First, the orientation angle $\psi$ is set to the desired one (dashed line) followed by motion to the desired position described in polar coordinates $(r,\phi)$.}
    \label{fig:traj_example}
\end{figure}
\begin{table}[]
\centering
\caption{Success rate for completing tasks with the VFM}
\label{tb:demo}
\begin{tabular}{lcccc}\toprule
                & Thickness  & Weight & Max. $r_c$ & Success rate \\
                & (mm) & (g) & (mm) & (\%) \\\midrule
Credit card     & 0.76 & 14  & 50 & 100\\
Ruler           & 0.60 & 17  & 20 & 70\\
Cellphone       & 7.85 & 172 & 75 & 90\\
\bottomrule
\end{tabular}
\end{table}

\begin{figure*}[]
    \includegraphics[width=\linewidth]{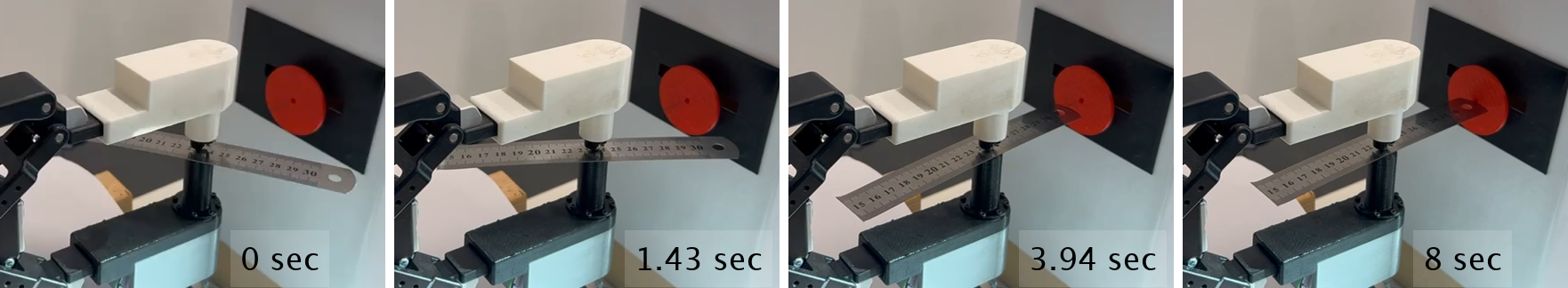} 
    \caption{Example trial of manipulating a steel ruler to touch a red circle.}
    \label{fig:ruler}
\end{figure*}
\begin{figure}[]
    \includegraphics[width=\linewidth]{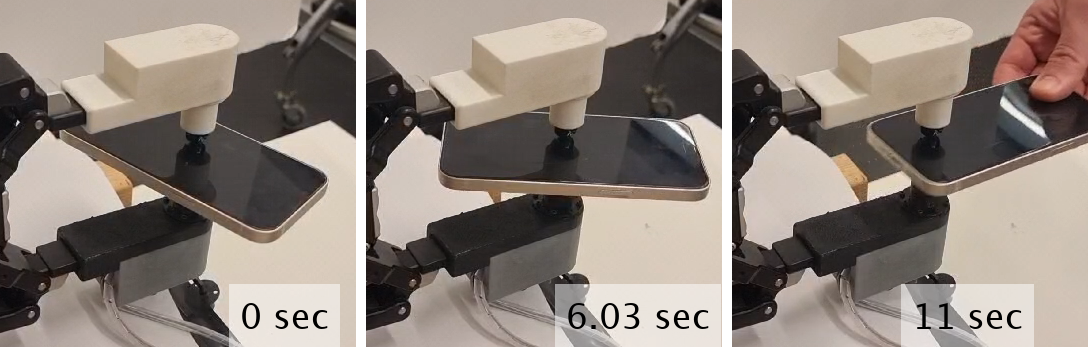} 
    \caption{Example trial of handing over a cellphone to a user.}
    \label{fig:cellphone}
\end{figure}

\subsection{Algorithm evaluation}

In this section, we evaluate the performance of Algorithm \ref{alg:main}. The duty cycle function $\Omega(t)$ in object translations was set to 50\%. Similarly, the object's rotation is conducted with vibration frequency of $\omega_o=168$ Hz. Furthermore, to initiate the cyclic motion, vibration with an excitation angle in the range $\theta_{\Delta t}\in[\frac{\pi}{6},\frac{\pi}{4}]$ was applied for $\Delta t=0.1sec$ before setting $\theta=\pi$ as presented in the algorithm. These values were calibrated manually based on trial and error, but are effective to all tested objects and further used in all experiments. In addition, the radius and orientation tolerance error bounds are $\epsilon_r= 1~mm$ and $\epsilon_\psi=1^\circ$, respectively.

We first demonstrate the cyclic motion with a disk 3D printed from PLA at a radius of 100 mm and thickness of 2 mm. Figure \ref{fig:circle} shows an example of a cyclic motion of radius 7.75 mm. Periodic variations of the $r$ and $\Psi$ values are seen. The cyclic motion was achieved 100\% of the 10 trials tested in this experiment.
Next, we evaluate the performance of the entire algorithm and its ability to drive the object to a desired state. The disk and a rectangular plate of size $80\times110$ mm were tested over 100 trials. In each trial, a random goal state (i.e., object position and orientation) was randomly sampled from within the possible workspace and from a uniform distribution, and the robot's motion was initiated from the endpoint of the previous trial. Then, the algorithm was initiated with first rotating the object to the desired orientation with the cyclic motion, followed by motion to the target position through the COM. We also compare the use of duty cycle function $\Omega(t)$ in translation over simple constant vibration frequency $\omega_o=240$ Hz.

Table \ref{tb:algorithm_accuracy} summarizes the mean position and orientation errors for both objects. First, the position errors match the values acquired in \cite{Nahum2022} and are approximately within the accuracy bounds of the measurement method. When observing the orientation errors, the ability of the duty cycle control to maintain the reached orientation angle is clearly observed. Without duty cycle excitation, the orientation of the object varies during the motion to the target position, even if passed through the origin. When operating the ERM with duty cycle excitation, the system is able to the drive the objects to the desired states with high accuracy. Figures \ref{fig:traj_snapshots} and \ref{fig:traj_example} present snapshots and state response, respectively, of one trial moving to a target state, with position and orientation errors of 0.76 mm and 2$^\circ$, respectively.


\subsection{Task demonstrations}

We next demonstrate the ability of the VFM to manipulate various objects in specific tasks. Three tasks are chosen: manipulating a credit card into an ATM-like slot; driving a ruler to touch a circle on a vertical wall; and handing over a cellphone to a human's hand. Each of these objects have different thickness, weight, surface texture and maximum possible rotation radius $r_c$, presenting a diverse set of challenges for the manipulation mechanism. For each task, 10 trials were run such that the goal states were predefined and initial states were randomly selected in each trial. Table \ref{tb:demo} presents the success rates for the three tasks. The table also presents the thicknesses and weights of the objects, and their maximum possible radii $r_c$ derived from their widths. The results show high success in completing the tasks. For the ruler, the maximum $r_c$ is low due to its narrow shape, increasing the probability of dropping the object during the cyclic motion. While the success rate is relatively high, one can cope with this difficulty by having precise control on the grasp force, yielding controlled damping. The results also demonstrate the ability to manipulate a thick and quite heavy object such as a cellphone. Trial examples for the credit card, ruler and cellphone are seen in Figure \ref{fig:front}, \ref{fig:ruler} and \ref{fig:cellphone}, respectively.

%% file: conclusions.tex
In this letter, we have proposed a novel algorithm to control the full state of a thin object, grasped by a parallel gripper, while using a vibration-based mechanism. The mechanism enhances the manipulation capabilities of parallel grippers, which on their own are incapable of performing complex manipulations. Using a comprehensive analysis of the system, we were able to propose an algorithm to manipulate the object to desired position and orientation. Additionally, the use of duty cycle was included to improve orientation accuracy. 
A finite element model of the dynamic system has been presented for validation the mathematical model and for parametric investigation of factors effecting the motion of the object such as 
vibration frequency. An experimental setup was tested on a variety of objects and demonstrated high precision. Furthermore, functional experiments were conducted, such as inserting a credit card into an ATM, aligning a ruler toward a specific target, and rotating a mobile device toward a user. These experiments exhibited high success rates.

As discussed previously, the system is underactuated, meaning it is not feasible to simultaneously control both the position and orientation of the object. Specifically, orientation control must be executed first, followed by position control. Therefore, future work may consider an improved design having two opposing vibration fingers. Such a configuration would enable simultaneous control of the entire object's state, allowing for enhanced capabilities in path and trajectory tracking. Current limitations, such as reliance on fiducial markers and camera line-of-sight, suggest that future work could explore the integration of on-board sensing modules, combining visual perception and odometry. Furthermore, improved calibration of the finite element model can enhance the control algorithm's performance and enable more accurate predictions for different cases. Additionally, expanding the scope to three-dimensional manipulation through vibration-based robotic hands could be a promising direction.